\documentclass{article}

    \PassOptionsToPackage{numbers, compress}{natbib}
    \usepackage[preprint]{neurips_2025}

\usepackage[utf8]{inputenc} % allow utf-8 input
\usepackage[T1]{fontenc}    % use 8-bit T1 fonts
\usepackage{url}            % simple URL typesetting
\usepackage{booktabs}       % professional-quality tables
\usepackage{amsfonts}       % blackboard math symbols
\usepackage{nicefrac}       % compact symbols for 1/2, etc.
\usepackage{microtype}      % microtypography
\usepackage{xcolor}         % colors
\usepackage{hyperref}       % hyperlinks

\definecolor{link}{RGB}{225,0,0}
\definecolor{cite}{RGB}{218,112,214}
\hypersetup{
    colorlinks=true,
    linkcolor=link,
    citecolor=cite}
\usepackage{url}

\usepackage{amsmath}
\usepackage{caption}
\usepackage{graphicx}
\usepackage[caption=false, font=footnotesize]{subfig}
\usepackage[capitalize]{cleveref}
\Crefname{section}{Sec.}{Secs.}
\Crefname{section}{Section}{Sections}
\Crefname{table}{Table}{Tables}
\Crefname{table}{Tab.}{Tabs.}
\usepackage{algorithm}
\usepackage{algorithmic}
\usepackage{wrapfig}
\usepackage{picinpar}
\usepackage{cutwin}
\usepackage{makecell}
\usepackage{array}
\usepackage{color}
\usepackage{colortbl}
\usepackage{multirow}
\usepackage{multicol}
\usepackage{comment}
\usepackage{amssymb}
\usepackage{marvosym}
\usepackage{arydshln}
\usepackage{subcaption}

\usepackage{pifont}

\definecolor{my_green}{RGB}{51,102,0}
\definecolor{my_yellow}{RGB}{255,165,0}
\definecolor{my_red}{RGB}{204, 0, 0}
\newcommand{\colorcmark}{\textcolor{my_green}{\ding{52}}}
\newcommand{\colorxmark}{\textcolor{my_red}{\ding{55}}}

\title{HaploOmni: Unified Single Transformer for Multimodal Video Understanding and Generation}

\author{%
  Yicheng Xiao$^{1,2}$\thanks{Equal contribution. $\textsuperscript{\Letter}$ Corresponding author.}\ \ , 
   Lin Song$^{2*}\textsuperscript{\Letter}$,
  \textbf{Rui Yang$^{3}$,}
  \textbf{Cheng Cheng$^{4}$,}
  \textbf{Zunnan Xu$^{1}$,}\\
  \textbf{Zhaoyang Zhang$^{2}$,}
  \textbf{Yixiao Ge$^{2}$,}
  \textbf{Xiu Li$^{1}\textsuperscript{\Letter}$,}
  \textbf{Ying Shan$^{2}$} \\
  \\$^{1}$Tsinghua Shenzhen International Graduate School, Tsinghua University
  \\$^{2}$ARC Lab, Tencent PCG \\ $^{3}$The University of Hong Kong \quad $^{4}$Xi’an JiaoTong University \\
  \texttt{xiaoyc23@mails.tsinghua.edu.cn \quad ronnysong@tencent.com}
}

\begin{document}

\maketitle

\begin{abstract}
With the advancement of language models, unified multimodal understanding and generation have made significant strides, with model architectures evolving from separated components to unified single-model frameworks.
This paper explores an efficient training paradigm to build a single transformer for unified multimodal understanding and generation.
Specifically, we propose a multimodal warmup strategy utilizing prior knowledge to extend capabilities.
To address cross-modal compatibility challenges, we introduce feature pre-scaling and multimodal AdaLN techniques.
Integrating the proposed technologies, we present the HaploOmni, a new single multimodal transformer.
With limited training costs, HaploOmni achieves competitive performance across multiple image and video understanding and generation benchmarks over advanced unified models. All codes will be made public at \href{https://github.com/Tencent/HaploVLM}{https://github.com/Tencent/HaploVLM}.
\end{abstract}

\section{Introduction}
In recent years, large-scale language models (LLMs)~\cite{llama3, qwen2, gpt4} have exhibited remarkable capabilities across diverse domains, prompting researchers to extensively investigate their potential applications in multimodal contexts.
There is an increasing focus on developing unified approaches that simultaneously address both multimodal understanding and generation capabilities. The former research can be categorized into three phases in terms of implementation architecture, progressing from segregated to unified frameworks.

In the first phase, tool-based methods like InstructGPT~\cite{instructgpt} and HuggingGPT~\cite{huggingpt} employ LLMs to allocate task-specific tools. While these methods offer simplicity and ease of use, their reliance on text-tool interactions limits their flexibility and controllability. In the second phase, methodologies incorporate separate encoders and decoders in conjunction with LLMs, exemplified by Seed~\cite{seedx}, Emu-2~\cite{emu2}, and VILA-U~\cite{vilau}, achieving multimodal input-output compatibility through feature interaction mechanisms. Although these approaches have achieved commendable results on general multimodal benchmarks, their segregated processes result in insufficient modal integration, constraining their capability to handle fine-grained understanding and generation tasks.

\begin{table*}[h]
    \centering
    \footnotesize
    \renewcommand\arraystretch{1.0}
    \setlength{\tabcolsep}{5.7pt}
    \begin{tabular}{lcccccccc}
    \toprule
    \multirow{2}{*}{Method} & Video & Single & Und. & Gen. & \multirow{2}{*}{SEED} & \multirow{2}{*}{POPE} & \multirow{2}{*}{MVBench} & \multirow{2}{*}{VBench}\\
    & Support & Transformer & Data & Data &  &  &  & \\
    \midrule
    SEED-X~\cite{seedx} & \colorxmark & \colorxmark & 152M  & 152M & - & 84.2 & - & - \\
    TokenFlow~\cite{tokenflow}  & \colorxmark & \colorxmark & 10M  & 60M & 68.7 & 86.8 & - & - \\
    Janus-Pro~\cite{januspro} & \colorxmark & \colorxmark & 41M  & 98M & 72.1 & 87.4 & - & - \\
    Show-o~\cite{showo}  & \colorxmark & \colorcmark & 36M  & 611M & - &  73.8 & - & -\\
    ViLA-U~\cite{vilau} & \colorcmark & \colorxmark & 7M  & 16M  & 59.0 & 85.8 & 38.9 & 73.4 \\
    \rowcolor{gray!10} HaploOmni (ours) & \colorcmark & \colorcmark & \textbf{4M} & \textbf{3M} & \textbf{74.6} & \textbf{88.3} & \textbf{52.9} & \textbf{78.1} \\
    \bottomrule
    
    \end{tabular}
    \caption{Characteristics comparison with some other unified models. Video support means that the models can process video inputs and generate videos. ``Und. Data” and ``Gen. Data” indicate the number of training data for understanding and generation tasks, respectively.}
    \label{tab:characteristics}
    % \vspace{-10pt}
\end{table*}

In the third phase, the latest approaches utilize a unified single-transformer framework. One subset, including Chameleon~\cite{chameleon} and Show-o~\cite{showo}, achieves model unification through image discretization tokens.
Another subset, exemplified by Transfusion~\cite{transfusion}, employs hybrid text autoregressive and image diffusion modeling processes for unification.
Compared to the encoder-decoder methods, these single-transformer methods are more streamlined and enable cross-modal early-fusion and late-fusion, thereby enhancing fine-grained multimodal representation capabilities~\cite{transfusion}.
However, existing methods adopt from-scratch training approaches.
Due to the absence of prior knowledge, their overall performance falls short of encoder-decoder methods while incurring substantial training costs.
Consequently, this paper explores a new perspective: \textit{efficiently constructing a single multimodal transformer by leveraging knowledge from specialized models to achieve high-performance unified multimodal understanding and generation.}

To achieve it, we propose a new training paradigm for single multimodal transformers.
Considering that the natural language possesses more abstract and higher-level semantic representations compared to natural images~\cite{llava}, we propose a multimodal warmup process that depth-wise partitions a transformer decoder into three components: visual encoding, text encoding-decoding, and visual decoding.
These components are initialized using corresponding prior models and subsequently fine-tuned independently to accommodate identity mapping across other modalities.
Following the warmup phase, the model undergoes unified training for multimodal understanding and generation in an end-to-end manner.
Furthermore, we find that different modalities exhibit varying preferences for feature scaling, significantly impacting training effectiveness and stability.
Inspired by the diffusion transformer, we propose feature pre-scaling strategies and Multimodal AdaLN.
The former pre-establishes initial feature transformation scales for different modalities based on statistical information, while the latter enables the model to autonomously select normalization parameters for various inputs.

With the proposed techniques, we present the \textbf{HaploOmni}, a cost-efficient yet high-performance single transformer for multimodal understanding and generation.
As demonstrated in~\cref{tab:characteristics}, we evaluate our method on image and video multimodal understanding and generation benchmarks.
Compared with previous models, our HaploOmni achieves superior performance across multiple image understanding datasets, including SEEDBench~\cite{li2023seed} and POPE~\cite{pope}.
Additionally, it significantly outperforms unified video-text models such as VILA-U in both MVBench~\cite{mvbench} video understanding and VBench~\cite{vbench} generation benchmarks.

\section{Related Work}
\label{related_work}

\noindent \textbf{Text-to-video generation models} aim to automatically produce visually and logically consistent videos based on textual descriptions of scenes, objects, and actions. Most text-to-video models~\cite{ho2022video, cogvideo, he2022latent, videocrafter} are built on latent diffusion models with a U-Net architecture. The field achieved a significant milestone with the introduction of diffusion transformers~\cite{DiT}, as demonstrated by the impressive Sora~\cite{sora}. Following this breakthrough, the majority of studies have adopted diffusion transformers to develop open-source text-to-video models. For example, CogVideoX~\cite{cogvideox} introduces an expert transformer to improve the fusion of visual and textual modalities. 

\noindent \textbf{Unified multi-modal LLMs} are capable of performing both understanding and generation tasks within a single framework. Several efforts~\cite{seedx, vilau, emu3, mindomni} have been made to unify vision understanding and generation. For instance, SEED~\cite{seedx} and Emu~\cite{emu2} predict the next multimodal element by regressing visual embeddings or classifying textual tokens. These models primarily rely on an additional diffusion-based image decoder~\cite{sdxl}. Similarly, Chameleon~\cite{chameleon} and Emu3~\cite{emu3} discretize visual features and train token-based autoregressive models on mixed image and text data. VILA-U~\cite{vilau} improves the vision tokenizer by introducing a unified vision tower that aligns discrete visual features with text during pre-training. In addition, TransFusion~\cite{transfusion} and Show-o~\cite{showo} attempt to integrate diffusion and autoregressive approaches within a single transformer. However, most unified models still lag behind task-specific architectures, likely because generation tasks require low-level features while understanding tasks demand high-level features. To address this limitation, Janus~\cite{janus} employs separate tokenizers for understanding and generation tasks. Similarly, TokenFlow~\cite{tokenflow} defines dual codebooks with shared mappings to enable flexible combinations of low and high-level features.
Despite recent advances, current approaches are limited by their inability to effectively trade off between performance and training resources.
In this paper, we introduce a method for efficiently constructing a unified single transformer achieving comparable performance across both understanding and generation tasks.

\begin{figure*}
    \centering
    \includegraphics[width=\linewidth]{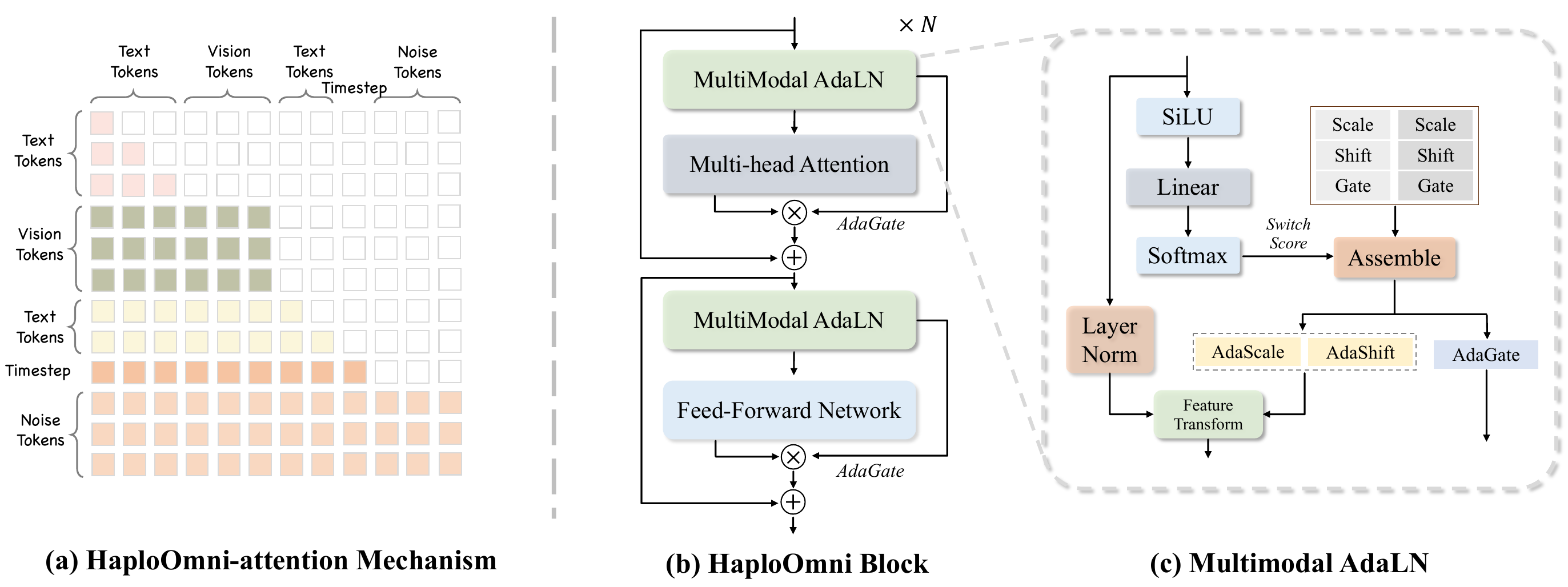}
    \caption{Illustration of our HaploOmni-attention mechanism and HaploOmni Block. We implement a hybrid masking strategy that applies causal attention to text features and timestep tokens while adopting bidirectional attention for processing visual signals and latent noise. Drawing from the standard transformer module, we develop the HaploOmni block through the implementation of multimodal AdaLN.}
    \label{fig:module}
\vspace{-3ex}
\end{figure*}

\section{Method}
In this section, we begin by introducing the preliminaries, followed by a detailed elaboration of our unified single transformer (HaploOmni) and the novel training paradigm we propose.
This approach leverages knowledge from specialized models to efficiently construct HaploOmni, enabling high-performance unified multimodal understanding and generation.

\subsection{Preliminaries}

\paragraph{Multimodal LLMs.}
Given a visual signal (image/video) and a series of corresponding text requests, a common approach for answer generation is to use a multimodal large language model~\cite{llava, qwen2, internvl}, which typically integrates a vision encoder and a language model.
Generally, the raw visual input is transformed into a discrete or continuous feature space, which is then combined with text embeddings generated by a linguistic tokenizer.
An auto-regressive LLM then processes the mixed multimodal sequence $\{x_t\}_{t=1}^{T-1}$ to predict the next tokens by modeling the conditional probability:
\begin{equation}
    P\left(x_{1}, x_{2}, \cdots, x_{T}\right)=\prod_{t=1}^{T} P\left(x_{t} \mid x_{1}, x_{2}, \cdots, x_{t-1}\right).
\end{equation}
Then, the NTP loss is defined using cross-entropy and the conditional probability described above, utilized to optimize the LLM during the training phase.

\begin{figure*}
    \centering
    \includegraphics[width=\linewidth]{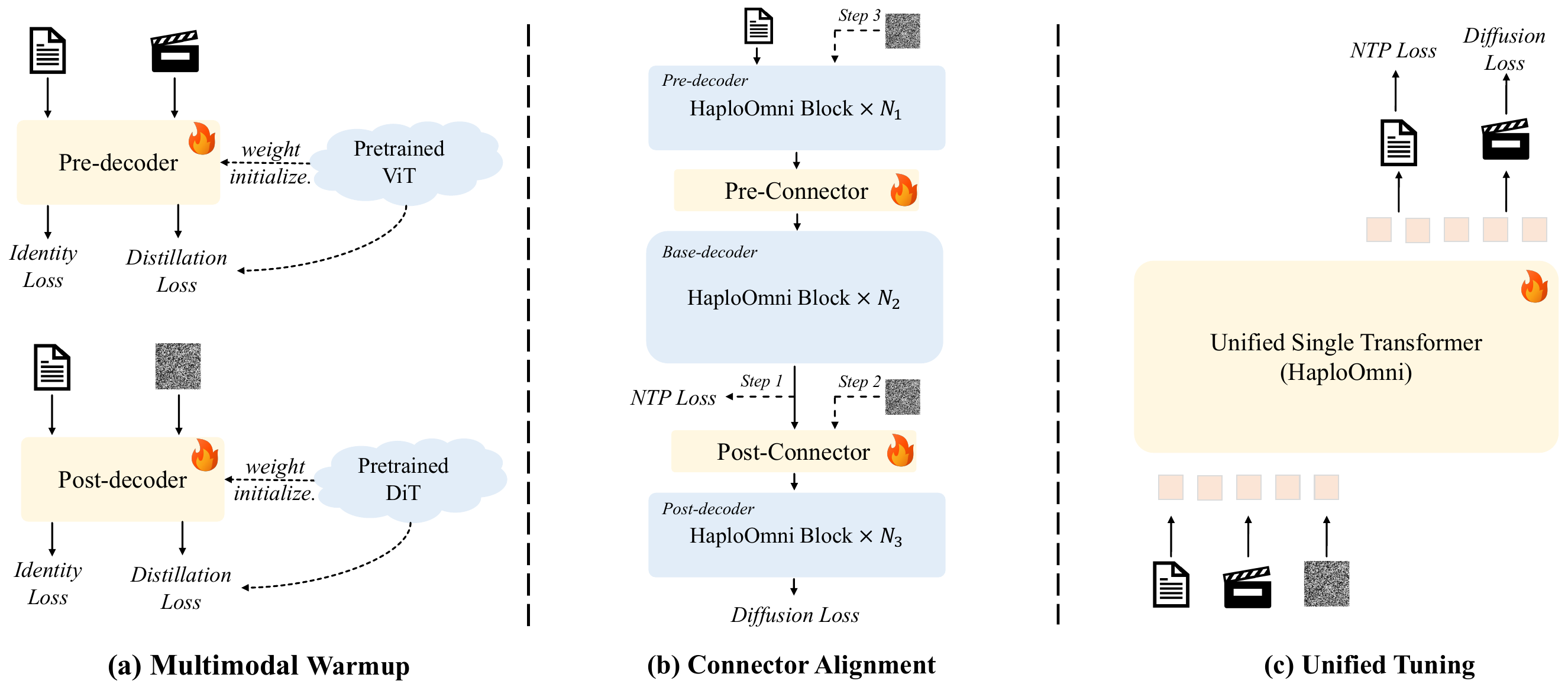}
    \caption{The progressive training stages of our HaploOmni, including multimodal warmup, connector alignment and unified tuning.}
    \label{fig:framework}
\end{figure*}

\paragraph{Diffusion Transformer.}
Diffusion models, such as the denoising diffusion probabilistic model (DDPM), generate data by progressively transforming noise into a target distribution over a series of timesteps.
The Diffusion Transformer (DiT) integrates the transformer architecture into this generative process, enabling it to learn the reversal of the incremental noise-adding procedure in the forward process.
At each timestep $t$, the model estimates the noise $\epsilon_t$ added to the data at the previous timestep. The objective function for training the Diffusion Transformer can be written as:
\begin{equation}
\mathcal{L}_{\textit{diff}} = \mathbb{E} \left[ \| \epsilon_t - \hat{\epsilon}_\theta(\mathbf{x}_t, t) \|^2 \right]
\end{equation}
where $\mathbf{x}_t$ is the noisy data at timestep $t$, $\mathbf{x}_0$ is the raw image or video data, and  $\hat{\epsilon}_\theta(\cdot)$ is the model's estimate of the noise at each timestep, parameterized by the network $\theta$. 
The attention mechanism in the transformer architecture enables the model to refine the noisy data by conditioning on both the input noise and the context provided by previous timesteps.
This iterative refinement process allows the model to generate high-quality samples from noise in both the image and text domains, making it particularly suitable for multimodal generative tasks.
Moreover, it lays a solid foundation for adapting the pre-trained DiT model to the inference paradigm of LLM, enabling the efficient development of a unified transformer model for multimodal understanding and generation.

\subsection{Model Design}
Overall, to streamline the training of the unified single transformer for multimodal understanding and generation, we first partition it into three components: pre-decoder, base-decoder, and post-decoder.
All parts consist of multiple HaploOmni Blocks as shown in~\cref{fig:module} and then are initialized using corresponding prior models, ViT~\cite{vit} for visual encoding, a pre-trained LLM~\cite{llama3} for text encoding-decoding, and DiT~\cite{cogvideox} for visual decoding.
Following this, two connector modules are employed to integrate the above three components into a complete transformer decoder.
In contrast to previous decoupled paradigms~\cite{seedx, emu2, janus}, our unified architecture processes both visual and textual inputs together, eliminating the need for a separate vision encoder, and enabling direct end-to-end visual generation conditioned on multimodal instructions.
Additionally, by leveraging the prior knowledge from pre-trained models, we significantly reduce the training resources required.
The following section provides an in-depth explanation of the specific modules within our HaploOmni.
\paragraph{HaploOmni-attention Mechanism.}
In light of the distinct characteristics of visual and linguistic modalities, we adopt a HaploOmni-attention mechanism with an adaptive mask strategy as shown in~\cref{fig:module} (a) to improve multimodal representation capacity following previous methods~\cite{showo, transfusion, omnigen}.
Specifically, we deploy bidirectional attention to preserve the intrinsic nature of visual signals in the continuous state space during the multimodal understanding process.
In the multimodal generation period, a causal mask is applied for the timestep token but a bidirectional mask for noise tokens.
In both periods, causal attention is applied to text features to maintain the causal dependencies in language.
\paragraph{HaploOmni Block.}
First, inspired by the expert adaptive LayerNorm (AdaLN) introduced by CogVideoX to facilitate the fusion of different modalities by separately normalizing the condition and noise embeddings, we develop a multimodal AdaLN as shown in~\cref{fig:module}(c).
Considering that AdaLN breaks the internal coherence required for constructing a one-transformer model, we introduce a dynamic strategy for input-aware normalization.
Specifically, we compute the state matrix $S$ offline to store two sets of scale, shift, and gate parameters as follows:
\begin{equation}
\label{equa:adascale}
S=
\begin{bmatrix}
 \text{Scale}_{\text{cond}}  & \text{Scale}_{\text{noise}} \\
 \text{Shift}_{\text{cond}}  & \text{Shift}_{\text{noise}} \\
 \text{Gate}_{\text{cond}}  & \text{Gate}_{\text{noise}}
\end{bmatrix}
= \text{SiLU}(\theta) W_{\text{Ada}}^\top, 
\end{equation}
\begin{wrapfigure}{r}{0.5\textwidth}  % r: right, l: left
\vspace{-10pt}  % 可选：控制与前文间距
\begin{minipage}{0.5\textwidth}
\begin{algorithm}[H]
\small
    \caption{Multimodal AdaLN}
    \label{alg:Switchable_AdaLN}
    \textbf{Input:}
    \begin{algorithmic}
        \STATE $h_i\in \mathbb{R}^{1\times d}$ \hfill $\triangleright$ \textit{Input feature}
        \STATE $W_{\text{MAL}}\in \mathbb{R}^{d\times 2}$ \hfill $\triangleright$ \textit{Input learnable matrix}
        \STATE $S\in \mathbb{R}^{3\times 2}$ \hfill $\triangleright$ \textit{State matrix}
    \end{algorithmic}
    \textbf{Forward:}
    \begin{algorithmic}
            \STATE $\overline{h_i} \gets \frac{h_i}{1+exp(-h_i)} W_{\text{MAL}}^\top$ 
            \STATE \textbf{Set} $\delta \in \mathbb{R}^{1\times 2}$
            \STATE $\delta_{k}\gets \frac{exp(\overline{h_i^k})}{{\textstyle \sum_{j=1}^{2}}exp(\overline{h_i^j}))}$ \hfill $\triangleright$ \textit{Switch Score}
            \STATE $[\text{AdaScale}, \text{AdaShift}, \text{AdaGate}] \gets \delta S^\top$
            \STATE \textbf{Do:} \hfill $\triangleright$ \textit{Feature Transform}
            \STATE \quad$\widetilde{h_i} \gets (\text{AdaScale} + 1) \times LN(h_i) + \text{AdaShift}$ 
    \end{algorithmic}
    \textbf{Output:} $\widetilde{h_i}$, $\text{AdaGate}$
\end{algorithm}
\end{minipage}
\vspace{-10pt}  % 可选：控制与后文间距
\end{wrapfigure}
where SiLU, $\theta$, and $W_\text{Ada}$ are the activation function, frozen time embedding, and a learnable matrix, respectively.
Based on the input feature $h_i$ of the $i$-th token in the sequence, we compute two switch score sets used to perform a weighted summation over the discrete state matrix.
The resulting AdaScale, AdaShift, and AdaGate parameters are then applied in the following feature transformation.
The detailed operation is shown in~\cref{alg:Switchable_AdaLN}.
Leveraging the Multimodal AdaLN, we develop a HaploOmni block which is used to construct the complete model.
The block is derived from the standard transformer structure, which includes two normalization layers, a feed-forward network, and an attention layer, with its execution order and residual method adhering to the original design, as depicted in ~\cref{fig:module} (b).

\paragraph{Pre \& Post Connector.}
Integrating specific decoders leads to discrepancies in the feature space across different modalities, which poses challenges for joint training and modality fusion.
To alleviate this, we introduce a novel connector module with multimodal LN to align the modalities within a unified feature space.
Specifically, given a multimodal sequence $X$ with the length of $L$ concatenated by $\{x_1, x_2\}$ where $\{x_1, x_2\}$ indicates the condition tokens and latent noise tokens respectively, we utilize a set of LayerNorm with learnable transition matrices $W^{'}$ to process the sequence as follows:
\begin{equation}
\widetilde{X} = \text{SiLU}(\text{LayerNorm}(X))W^{'}
\end{equation}
Then, we obtain the corresponding switch scores $P^{\text{score}} \in \mathbb{R}^{L\times 2}$ through an indicator layer consisting of a SiLU function, a learnable matrix $W_{\text{SN}}$, and a Softmax function ($\sigma$), which can be formulated as:
\begin{equation}
P^{\text{score}} = \sigma(W_{\text{SN}}(\text{SiLU}({X})))
\end{equation}
With a characteristic function $\mathbb{I}$, the score is multiplied by the input $\widetilde{X}$ to obtain the well-aligned feature $\{X'_i\}_{i=1}^{L}$:
\begin{equation}
X'_i = \mathbb{I}_0(P^{\text{score}})\widetilde{X}_i + \mathbb{I}_1(P^{\text{score}})X_i 
\end{equation}

\paragraph{Feature Pre-scaling.}
Although the model can ultimately be optimized through the connector we designed, the optimization process is relatively slow.
We observe that aligning features across modalities gives rise to considerable amplitude inconsistencies, with the amplitudes of noise tokens often being about 10 times larger than those of the visual features distilled by a prior ViT.
This disparity intensifies feature-space distribution differences, complicating the training process.
Additionally, in our paradigm, small perturbations near extreme points, stemming from the pre-trained model, lead to diminished gradient amplitudes, which slow parameter updates.
Therefore, we introduce a feature pre-scaling mechanism into the Pre and Post-decoder, significantly simplifying training and accelerating model convergence.
\begin{table*}[t]
    \centering
    \setlength{\tabcolsep}{1.9pt}
    \renewcommand{\arraystretch}{1.2}
    % \scriptsize
    \scriptsize
    \begin{tabular}{llcccccccccc}
        \toprule
        \textbf{Type} & \textbf{Model} & \textbf{Size} & \textbf{SEED$ \uparrow$} & \textbf{POPE$ \uparrow$} & \textbf{AI2D$ \uparrow$} & \textbf{RWQA$ \uparrow$} & \textbf{MMMU$ \uparrow$} & \textbf{MMB$_{\text{(test)}}$$\uparrow$} & \textbf{MMStar$ \uparrow$} & \textbf{VQAv2$ \uparrow$} & \textbf{GQA$ \uparrow$} \\
        \midrule
        \multirow{13}{*}{\centering \textit{Und. Only}}
&MobileVLM-V2~\cite{mobilevlmv2} & 1.4B & - & 84.3 & - & - & - & 57.7 & - & - & 59.3 \\
&LLaVA-v1.5~\cite{llava_v1_5}& 7B & 66.1 & 85.9 & 54.8 & 54.8 & 35.3 & 64.3 & 30.3 & 78.5 & 62.0 \\
&InstructBLIP~\cite{InstructBLIP}& 7B & 58.8 & - & 33.8 & 37.4 & 30.6 & 36.0 & - & - & 49.2 \\
&Qwen-VL-Chat~\cite{qwenvl}& 7B & 58.2 & - & 45.9 & 49.3 & 35.9 & 60.6 & 37.5 & 78.2 & 57.5 \\
&InternVL-Chat~\cite{internvl}& 7B & - & 86.4 & 54.8 & - & - & - & - & 79.3 & 62.9 \\
&mPLUG-Owl2~\cite{mplug-owl2} & 7B & 57.8 & 86.2 & 55.7 & 50.3 & 32.7 & 64.5 & - & 79.4 & 56.1 \\
&ShareGPT4V~\cite{sharegpt4v}& 7B & - & - & 58.0 & 54.9 & 37.2 & 68.8 & 33.0 & 80.6 & 63.3 \\
&LLaVA-1.6~\cite{llava-next} & 7B & 64.7 & 86.5 & 66.6 & 57.8 & 35.1 & 67.4 & - & 81.8 & 64.2 \\
&VILA~\cite{vila}& 7B & 61.1 & 85.5 & - & - & - & 68.9 & - & 80.8 & 63.3 \\
&LLaVA-OV~\cite{llava_one_vision}& 7B & 75.4 & - & 81.4 & 66.3 & 48.8 & 80.8 & 61.7 & - & - \\
\cdashline{2-12}
\\[-2ex]
&Fuyu-8B~\cite{fuyu-8b}& 8B & - & 74.1 & 64.5 & - & 27.9 & 10.7 & - & 74.2 & - \\
&EVE-7B~\cite{eve}& 8B & 54.3 & 83.6 & - & - & - & 49.5 & 28.2 & 75.4 & 60.8\\
&Emu3-Chat~\cite{emu3}& 8B & 68.2 & 85.2 & 70.0 & 57.4 & 31.6 & 58.5 & - & 75.1 & 60.3 \\
\midrule
\multirow{10}{*}{\centering \textit{Und. and Gen.} }&LWM~\cite{LWM} & 7B & - & 75.2 & - & - & - & - & - & 55.8 & 44.8\\
&NExT-GPT~\cite{nextgpt}&13B & - & - & - & - & - & - & - & 66.7& - \\
&DreamLLM~\cite{dreamllm}& 7B & - & - & - & - & - & 58.2 & - & 72.9 & - \\
&VILA-U~\cite{vilau}& 7B & 59.0 & 85.8 & - & - & - & - & - & 79.4 & 60.8 \\
&Janus~\cite{janus}& 1.3B & 63.7 & 87.0 & - & - & 30.5 & 69.4 & - & 77.3 & 59.1 \\
&Janus-Pro~\cite{januspro}& 7B & 72.1 & 87.4 & - & - & 41.0 & 79.2 & - & - & 62.0 \\
\cdashline{2-12}
\\[-2ex]
&Chameleon~\cite{chameleon}& 30B & - & - & - & - & - & 37.6 & - & 69.6 & - \\
&Show-o~\cite{showo}& 1.3B & - & 73.8 & - & - & 25.1 & - & - & 59.3 & 48.7 \\
&TokenFlow-XL\cite{tokenflow}& 13B & 68.7 & 86.8& 66.7 & 53.7 & 38.7 & 68.9 & - & 77.9 & 62.7 \\
% &\cellcolor[HTML]{e6e6e6}\textbf{HaploOmni (ours)} & \cellcolor[HTML]{e6e6e6}10B & \cellcolor[HTML]{e6e6e6}74.8 &\cellcolor[HTML]{e6e6e6}87.9 &\cellcolor[HTML]{e6e6e6}76.6 &\cellcolor[HTML]{e6e6e6}60.8 &\cellcolor[HTML]{e6e6e6}39.7 &\cellcolor[HTML]{e6e6e6}73.5 &\cellcolor[HTML]{e6e6e6}55.2 &\cellcolor[HTML]{e6e6e6}76.7 &\cellcolor[HTML]{e6e6e6}62.6 \\
&\cellcolor[HTML]{e6e6e6}\textbf{HaploOmni (ours)} & \cellcolor[HTML]{e6e6e6}9B & \cellcolor[HTML]{e6e6e6}74.0 &\cellcolor[HTML]{e6e6e6}89.6 &\cellcolor[HTML]{e6e6e6}78.7 &\cellcolor[HTML]{e6e6e6}63.5 &\cellcolor[HTML]{e6e6e6}46.1 &\cellcolor[HTML]{e6e6e6}78.2 &\cellcolor[HTML]{e6e6e6}57.8 &\cellcolor[HTML]{e6e6e6}75.6 &\cellcolor[HTML]{e6e6e6}60.8 \\

        \bottomrule
    \end{tabular}
    \caption{Comparison with state-of-the-arts on image understanding benchmarks. ``Und.'' and ``Gen.'' denote ``understanding'' and ``generation'', respectively. Models below the dotted line are the single-transformer methods.}
    \label{image_und}
\vspace{-2ex}
\end{table*}
\paragraph{Inference Mode.}
In the inference stage, our model uses a unified transformer to execute multimodal understanding and generation tasks seamlessly.
For the understanding task, given a visual signal such as an image or video and a corresponding text query, the visual input is first converted into a sequence via a patchification layer, while the text is tokenized into a sequence. The concatenated multimodal sequences are then fed into the transformer and output with the corresponding response.
For the generation task, we combine condition tokens and random noise tokens into a multimodal sequence, process it iteratively through a unified transformer according to the DDIM~\cite{ddim} schedule, and decode the resulting latent representation into the final image or video using a VAE decoder~\cite{cogvideox}.

\subsection{Training Procedure}
HaploOmni is initially partitioned into three components: pre-decoder, base-decoder, and post-decoder.
We then train these components in three distinct stages: Multimodal Warmup, Connector Alignment, and Unified Tuning, as shown in ~\cref{fig:framework}.

\paragraph{Stage 1: Multimodal Warmup.}
The three sub-decoders are first initialized using the corresponding prior models and subsequently fine-tuned independently to accommodate identity mapping across other modalities.
At this stage, we only train pre-decoder and post-decoder to ensure they conform to the auto-regressive paradigm without altering the original model’s learnable parameters.
This adjHaploOmniment enables compatibility with the LLM reasoning framework, including KV-Cache, temperature setting, and top-p truncation.
For the pre-decoder, a mixed sequence of text and image tokens is used as input and we leverage the HaploOmni-attention mechanism for multimodal interaction.
Two losses are applied during training: Identity Loss for linguistic modality and distillation loss to preserve visual knowledge while learning new text-based knowledge.
On the other hand, we train the denoising capability of the post-decoder with randomly noisy video as input, conditioned by the corresponding text description, while applying both distillation loss and identity loss.

\paragraph{Stage 2: Connector Alignment.}
This stage aims to optimize the model training cycle across three progressive steps.
In the first step, the pre-connector is trained on multimodal understanding tasks with NTP loss.
In the second step, we train the post-connector, equipping the post-decoder to handle video and image denoising based on semantic features from the base LLM with diffusion loss.
Finally, we train both the pre-connector and post-decoder to allow the entire model to process visual, text, and latent noise features directly in an end-to-end manner.

\paragraph{Stage 3: Unified Tuning.}
At this stage, we integrate the three decoders into a unified single transformer (HaploOmni).
The entire model is fine-tuned using a combination of understanding and generation datasets.
Inputs across all modalities are uniformly processed through HaploOmni, which then generates the corresponding output. In this stage, we leverage both NTP loss and diffusion loss to optimize HaploOmni.

\begin{figure*}[t]
    \centering
    \includegraphics[width=\linewidth]{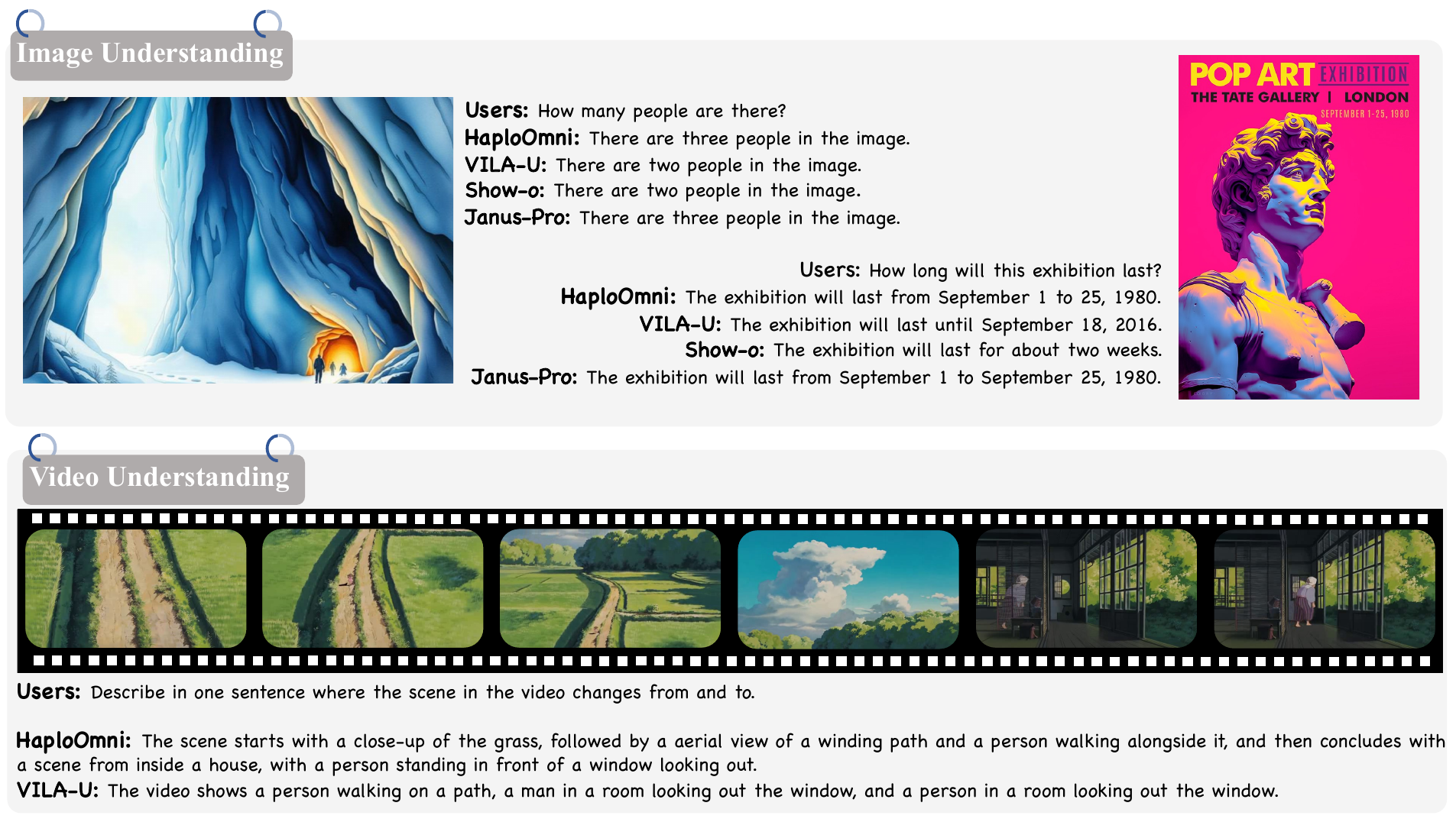}
    \vspace{-3ex}
    \caption{Performance comparison on image and video understanding capability.}
    \label{fig:com_und}
\vspace{-3ex}
\end{figure*}

\begin{figure*}[t]
    \centering
    \includegraphics[width=\linewidth]{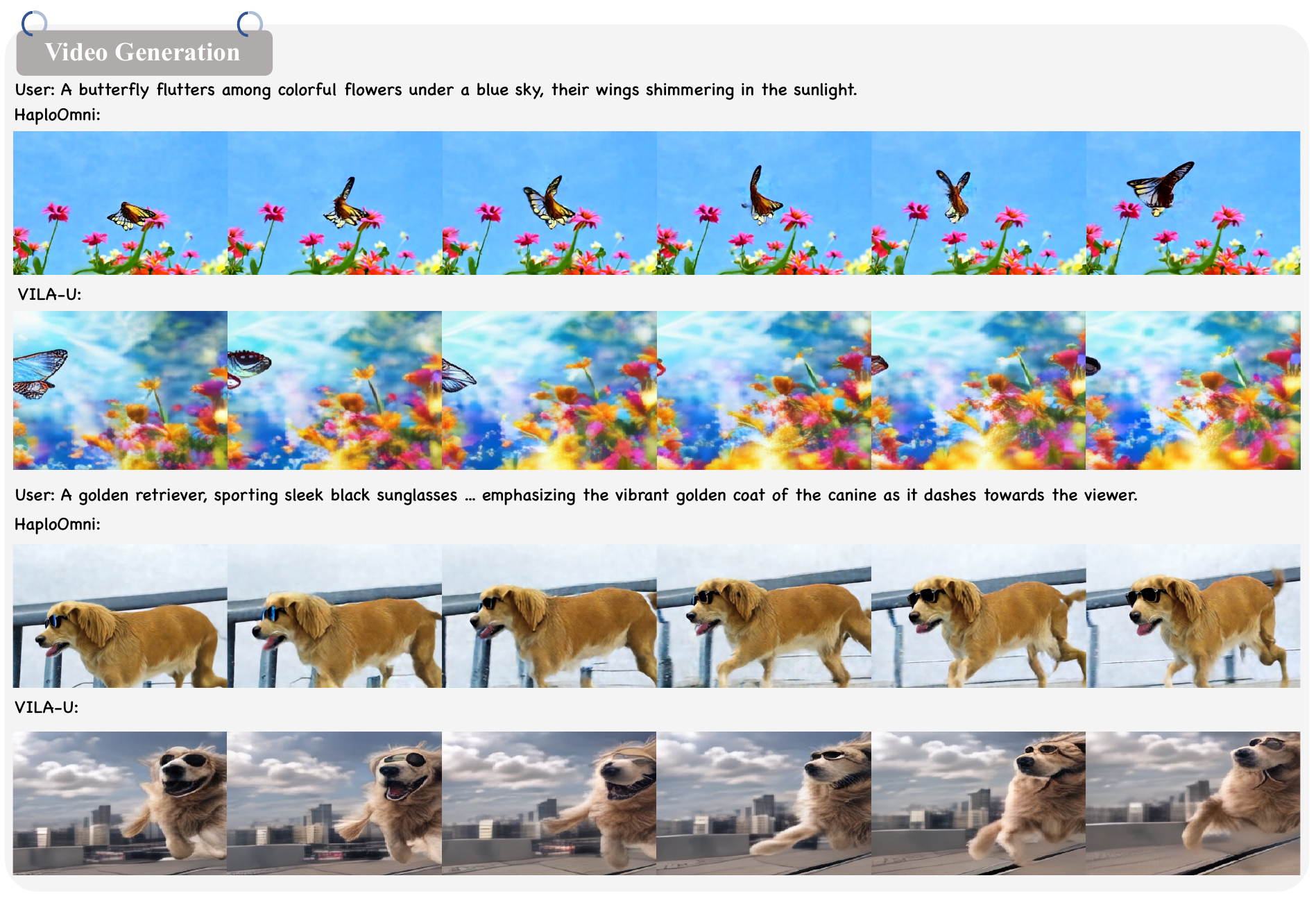}
    \vspace{-3ex}
    \caption{Performance comparison on video generation capabilities. The resolution of the generated video is 480 $\times$720.}
    \label{fig:com_gen}
\vspace{-4ex}
\end{figure*}
\section{Experiments}
\begin{table*}[ht]
    \centering
    \setlength{\tabcolsep}{5pt}
    \renewcommand{\arraystretch}{1.2}
    % \scriptsize
    \scriptsize
    \begin{tabular}{llccccc}
        \toprule
        \multirow{2}{*}{\textbf{Type}} & \multirow{2}{*}{\textbf{Model}} & \textbf{Subject} & \multirow{2}{*}{\textbf{Scene$ \uparrow$}} & \textbf{Dynamic} & \textbf{Motion} & \textbf{Background} \\
        &&\textbf{Consistency$ \uparrow$}&&\textbf{Degree$ \uparrow$}&\textbf{Smoothness$ \uparrow$} &\textbf{Consistency$ \uparrow$} \\
        \midrule
        \multirow{8}{*}{\centering \textit{Gen. Only}} & OpenSora-V1.1~\cite{open-sora-plan}& 96.8& 27.2& 47.7 & 98.3 &97.6\\
        &AnimateDiff-V2~\cite{animatediff}& 95.3& 50.2& 40.8 & 97.8&97.7\\
        &Pika~\cite{pika}& 96.9& 49.8 & 47.5 & 99.5&97.4\\
        &VideoCrafter-2.0~\cite{videocrafter}& 96.9& 55.3 & 42.5 & 97.7&98.2\\
        &CogVideoX-5B~\cite{cogvideox}& 96.2& 53.2 & 71.0 & 96.9&96.5\\
        &Kling~\cite{kling}&98.3&50.9&46.9&99.4&97.6\\
        &Gen-3~\cite{gen3}& 67.1&54.6&60.1& 99.2&96.6\\
        &Emu3-gen~\cite{emu3}& 95.3& 37.1&79.3&98.9&97.7\\
        \midrule
        \multirow{2}{*}{\centering \textit{Und. and Gen.}}&VILA-U~\cite{vilau}&87.0&31.8&58.7 & 95.3&94.4\\  
        &\cellcolor[HTML]{e6e6e6}\textbf{HaploOmni (ours)} &\cellcolor[HTML]{e6e6e6}96.4 &\cellcolor[HTML]{e6e6e6}34.6 &\cellcolor[HTML]{e6e6e6}65.3&\cellcolor[HTML]{e6e6e6}96.8&\cellcolor[HTML]{e6e6e6}97.6\\
        \bottomrule   
    \end{tabular}
    \caption{Comparison with state-of-the-arts on video generation benchmark, VBench~\cite{vbench}. ``Und.'' and ``Gen.'' denote ``understanding'' and ``generation'', respectively.}
    \label{video_gen}
\vspace{-2ex}
\end{table*}

\begin{figure}[h]
\centering
\begin{minipage}[c]{0.45\textwidth}
\centering
\includegraphics[width=\textwidth]{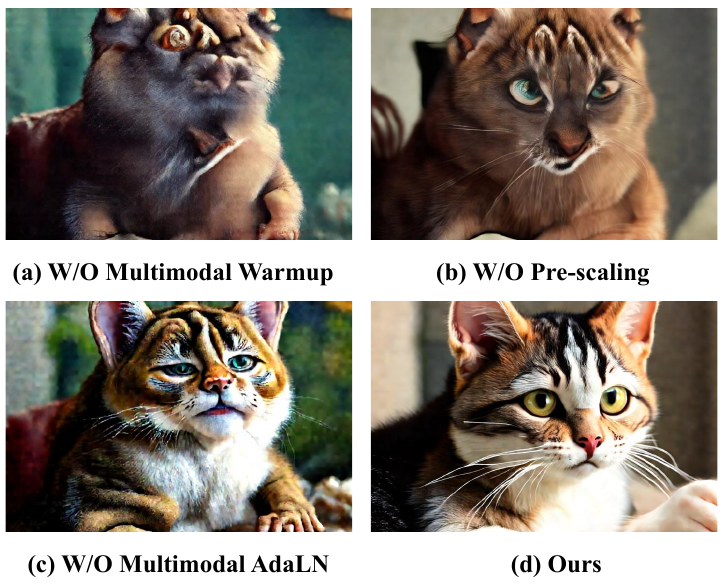}
\caption{Visualization results for the ablation of different components. (d) indicates the final version of our model.}
\label{fig:ablation_case}
\end{minipage}
\hspace{0.07\textwidth}
\begin{minipage}[c]{0.45\textwidth}
\centering
\includegraphics[width=0.93\textwidth]{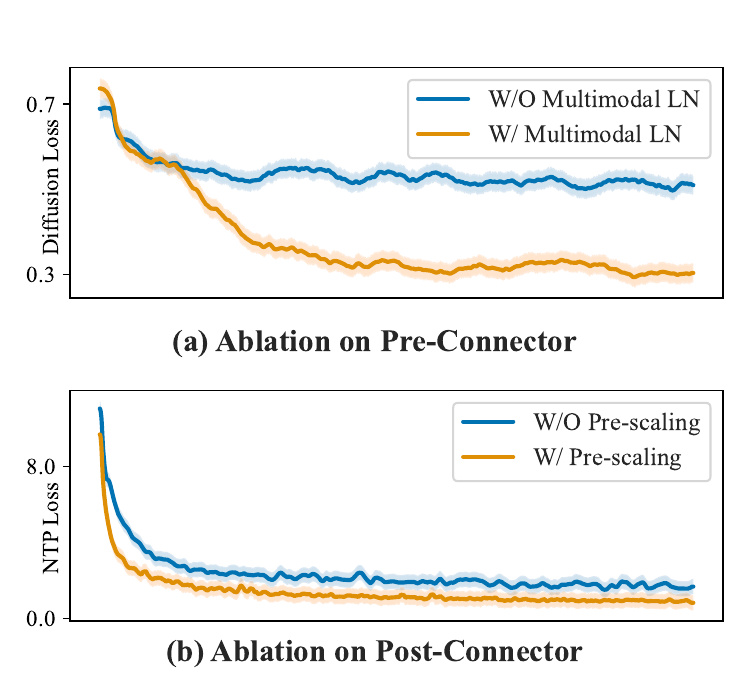}
% \vspace{0.1pt}
\caption{Loss curve comparison of different settings. The x-axis is the training step.}
\label{fig:loss_curve}
\end{minipage} 
\vspace{-7pt}
\end{figure}
We conduct extensive experiments to evaluate the effectiveness of our HaploOmni and compare it to the widely adopted large language model approaches on multimodal understanding and generation tasks under a fair evaluation setting.

\subsection{Datasets and Metrics}

\paragraph{Datasets.}
We classify image-text data pairs for multimodal understanding into three types consisting of 1.7M image caption data~\cite{sharegpt4v, llava}, 1.2M single-image instruction data~\cite{llava_v1_5, llava_one_vision} and 1.1M interleaved multi-image and video datasets~\cite{zhu2023minigpt, sharegptvideo, llava_one_vision}.
For the visual generation task, we curated 2M JourneyDB~\cite{journeydb} image-text pairs and approximately 1M video generation datasets, including 374K WebVid~\cite{webvid}, 626K in-house data.
More details are shown in the Appendix.

\paragraph{Metrics.}
For multimodal understanding, our model HaploOmni is evaluated on widely adopted image-based benchmarks.
For generation tasks, we evaluate our model on VBench~\cite{vbench}
, which involves various metrics such as dynamic degree, motion smoothness, and subject consistency.

\subsection{Implementation Details}
The base-decoder of our HaploOmni is based on Qwen2.5-7B~\cite{qwen2.5}.
During the distillation stage, we employ CLIP-ViT-L and CogVideoX-2B as the teacher models for the pre-decoder and post-decoder, respectively, with the decoders comprising 24 and 30 layers ($N_1$ and $N_2$).
Due to the limited space, more implementation details are shown in the Appendix.

\subsection{Main Results}
\paragraph{Visual Understanding.}
We provide a comparative analysis of state-of-the-art models on visual understanding across various benchmarks as depicted in~\cref{image_und} and~\cref{video_und} involving image and video, respectively.
As depicted in~\cref{image_und}, our HaploOmni, as a unified multimodal model outperforms existing methods on most evaluation metrics.
HaploOmni achieves state-of-the-art results among unified models on most benchmarks, with notable scores such as 74.8 on SEED and 87.9 on POPE, surpassing prior approaches like Janus and VILA-U.
Additionally, HaploOmni demonstrates competitive performance compared to understanding-only models, achieving scores of 76.6 on AI2D and 60.8 on RWQA, outperforming Emu3-chat by +6.6\% and +3.4\%, respectively.
Furthermore, the comparison results in \cref{video_und} highlight HaploOmni's impressive video understanding capabilities. Specifically, HaploOmni achieves 47.1 on EgoSchema and 52.9 on MVBench, surpassing Video-LaVIT with 37.3 on EgoSchema, and VILA-U with 38.9 on MVBench.
\begin{figure}[t]
  \centering
  \begin{minipage}[c]{0.47\textwidth}
    \centering
    \setlength{\tabcolsep}{2.1pt}
    \renewcommand{\arraystretch}{1.2}
    \footnotesize
    % \scriptsize
    \begin{tabular}{lccc}
      \toprule
      \textbf{Type} & \textbf{MMMU-val} & \textbf{MMStar} & \textbf{AI2D} \\
      \midrule
      Standard & 34.4 & 68.1 & 72.3\\
      HaploOmni-Block & 39.7 & 73.4 & 76.6 \\
      \bottomrule
    \end{tabular}
    \captionof{table}{Effectiveness of HaploOmni Block. A standard block refers to a commonly used block architecture in large language models (LLMs).}
    \label{tab:abl_block}
    \vspace{1.1cm}
    \setlength{\tabcolsep}{2.3pt}
    \renewcommand{\arraystretch}{1.2}
    \begin{tabular}{lccc}
      \toprule
       & \textbf{Chameleon} & \textbf{Janus} & \textbf{HaploOmni} \\
      \midrule
      Support Video & \colorxmark & \colorxmark & \colorcmark\\
      GPUs Hours & 856481 & 21504 & 5792\\
      \bottomrule
    \end{tabular}
    \captionof{table}{Comparison of training GPUs hours among some unified multi-modal large language models.}
  \end{minipage}
  \hfill
  % 右侧：单独横向的大表格 C
  \begin{minipage}[c]{0.5\textwidth}
    \centering
    \setlength{\tabcolsep}{1.6pt}
    \renewcommand{\arraystretch}{1.2}
    \footnotesize
    % \scriptsize
    \begin{tabular}{lccc}
        \toprule
        \textbf{Model} & \textbf{Size} &\textbf{EgoSchema$ \uparrow$} & \textbf{MVBench$ \uparrow$} \\
        \midrule
        \multicolumn{4}{c}{\textit{Und. only}} \\
        \midrule
        LLaMA-VID~\cite{llama-vid}&7B &38.5& 41.9\\
        Video-LLaVA~\cite{videollava}&7B& 38.4& 41.0\\
        VideoChat2~\cite{mvbench}&7B& 42.2& 51.1\\
        LLaVA-NeXT~\cite{llava-next}&7B& 43.9& 46.5\\
        LLaVA-OneVision&72B& 62.0&-\\
        VideoLLaMA2&7B& 51.7& 54.6\\
      \midrule
        \multicolumn{4}{c}{\textit{Und. and Gen.}} \\
        \midrule
        Video-LaVIT~\cite{videolavit}&7B&37.3&-\\
        VILA-U~\cite{vilau}&7B&33.4&38.9\\  
        \cellcolor[HTML]{e6e6e6}\textbf{HaploOmni (ours)} &\cellcolor[HTML]{e6e6e6}9B&\cellcolor[HTML]{e6e6e6}47.1 &\cellcolor[HTML]{e6e6e6}52.9\\
        \bottomrule   
    \end{tabular}
    \captionof{table}{Performance comparison on video understanding benchmarks. “Und.” and “Gen.” denote “understanding” and “generation”, respectively.}
    \label{video_und}
  \end{minipage}
\vspace{-7pt}
\end{figure}

\begin{figure*}[t]
    \centering
    \includegraphics[width=\linewidth]{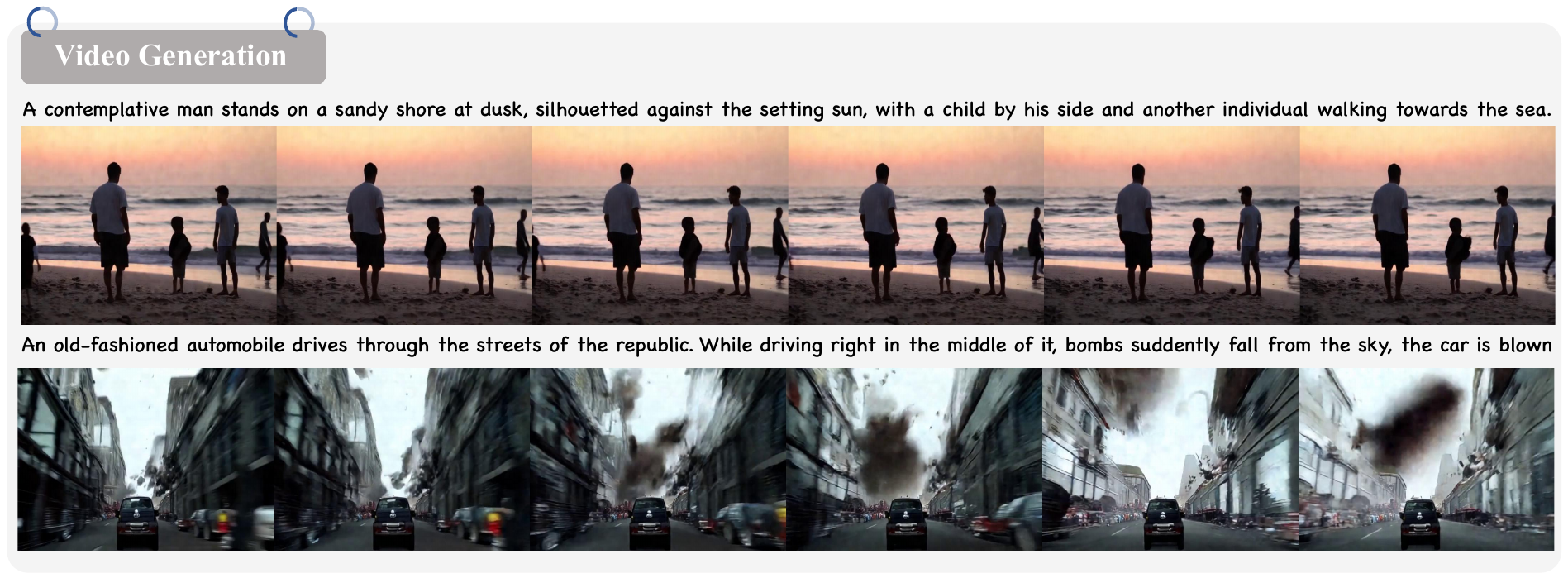}
    \vspace{-3ex}
    \caption{Qualitative results of HaploOmni. The resolution of all the generated videos is 480 ×720.}
    \label{fig:exp_case}
\vspace{-1ex}
\end{figure*}
\paragraph{Video Generation.}
We compare the performance of our proposed HaploOmni with state-of-the-art video generation models on the VBench benchmark as shown in~\cref{video_gen}.
Following previous works~\cite{emu3, cogvideox} on video generation, we selected some aspects that can reflect the quality of the generated video, like dynamic degree, subject consistency, and motion smoothness.
HaploOmni as a unified model, exhibits strong performance across most evaluated aspects. Specifically, we achieve a Scene Consistency score of 96.4, outperforming other multimodal models like VILA-U (87.0) while remaining competitive with pure generative models such as Kling (98.3) and Pika (96.9).
\subsection{Ablation Study}
We conduct various analysis experiments and present some visual results to illustrate the effectiveness of our method.
As shown in~\cref{fig:loss_curve}, multimodal LN effectively reduces the difficulty of visual generation training, while feature pre-scaling accelerates the training process for multimodal understanding and improves loss convergence.
We ablate various strategies by generating a cute cat as illustrated in~\cref{fig:ablation_case}.
Noise increases when multimodal AdaLN is absent.
Feature pre-scaling contributes to more accurate semantic tracking.
The comparison between (a) and (d) underscores the advantage of the warmup process.
As shown in \cref{tab:abl_block}, our HaploOmni Block outperforms the standard version under a fair evaluation protocol, which demonstrates the effectiveness of our architectural design.

\subsection{Qualitative Results}
To better illustrate the capabilities of our HaploOmni, we provide examples of image understanding, video understanding, and video generation.
As shown in~\cref{fig:com_und}, with the decoder-only architecture, the model can handle input images of varying resolutions and perceive the fine-grained information.
Meanwhile, HaploOmni effectively displays the motion range of generated concepts, such as the butterfly in~\cref{fig:com_gen} and building fragments in~\cref{fig:exp_case}.
More qualitative results are shown in~\cref{fig:more_case} of Appendix.

\section{Conclusion}
This paper explores a new training paradigm for single multimodal transformers. By introducing a multimodal warmup strategy incorporating prior knowledge, we substantially reduce training complexity and computational costs. Furthermore, we propose the feature pre-scaling strategy and multimodal AdaLN to address cross-modal integration challenges. With these techniques, our proposed HaploOmni demonstrates high performance in both image and video understanding and generation, achieving state-of-the-art results across multiple benchmarks. Additionally, we believe our methodological approach can inspire future LLM-based research.

\appendix
\section{Appendix}

\subsection{Datasets.}
We classify image-text data pairs for multimodal understanding into three types: 1) image caption data, which include 1.2M ShareGPT4V-PT~\cite{sharegpt4v} and 558K LLaVA pretraining data~\cite{llava}; 2) single-image instruction data, comprising 665K LLaVA v1.5~\cite{llava_v1_5} and 0.5M public dataset~\cite{llava_one_vision}; and 3) interleaved multi-image and video datasets, which consist of 0.6M CC3M~\cite{zhu2023minigpt}, LLaVA-Hound mixed data, and 0.5M video datasets~\cite{sharegptvideo, llava_one_vision}.
Furthermore, we follow existing works~\cite{janus, showo, haplovlm} to organize the above caption data into question-answering pairs.
For the visual generation task, we curated 2M JourneyDB~\cite{journeydb} image-text pairs and approximately 1M video generation datasets, including 374K WebVid~\cite{webvid}, 626K in-house data.

\subsection{Metrics.}
In multimodal understanding, our model HaploOmni is evaluated on widely adopted image-based benchmarks.
including GQA~\cite{gqa}, VQAv2~\cite{vqa}, AI2D~\cite{ai2d}, MMBench-EN-dev (MMB), MMMU~\cite{mmmu}, RealWorldQA, MMStar~\cite{mmstar}, POPE~\cite{pope} and SEED-Bench-IMG (SEED)~\cite{seedbench} as well as the video benchmarks, including MVbench~\cite{mvbench} and EgoSchema~\cite{egoschema}. 
For generation tasks, we evaluate our model on VBench~\cite{vbench}
, which involves various metrics such as dynamic degree, motion smoothness, and subject consistency.

\subsection{Implementation.}
The base-decoder of our HaploOmni is based on Qwen2.5~\cite{qwen2.5}.
During the distillation stage, we employ CLIP-ViT-L and CogVideoX-2B as the teacher models for the pre-decoder and post-decoder, respectively, with the decoders comprising 24 and 30 layers ($N_1$ and $N_2$).
In the decoder warmup stage, the pre-decoder is trained with a learning rate of 1e-4 and a batch size of 256, while the post-decoder is trained using a learning rate of 2e-4 and a batch size of 32.
In step 1 of the alignment stage, we align the pre-decoder and mid-decoder with a learning rate of 1e-5 and a batch size of 128, training only the pre-connector with a 2K-step warmup.
In step 2, the pre-connector is warmed up for 10K iterations using JourneyDB data with a learning rate of 1e-4 and a batch size of 128, after which we relax the training for the post-decoder.
In step 3, we train the pre-connector, post-connector, and post-decoder with the same settings, enabling end-to-end input-output of latent features.
Finally, in the third stage, the HaploOmni is fine-tuned uniformly with mixed video and image generation, as well as multimodal understanding data with a learning rate of 2e-5 and a batch size of 32. Across all experiments, the AdamW optimizer is configured with betas (0.9, 0.999) and a momentum of 0.9~\cite{adamw, mambatree, uvcom}.
By default, the number of multimodal AdaLN layers is set to 2.

\subsection{Broader Impact}
With these techniques, our proposed HaploOmni demonstrates high performance in both image and video understanding and generation, achieving state-of-the-art results across multiple
benchmarks.
Additionally, we believe our methodological approach can inspire future LLM-based research.
\begin{figure*}[ht]
    \centering
    \includegraphics[width=\linewidth]{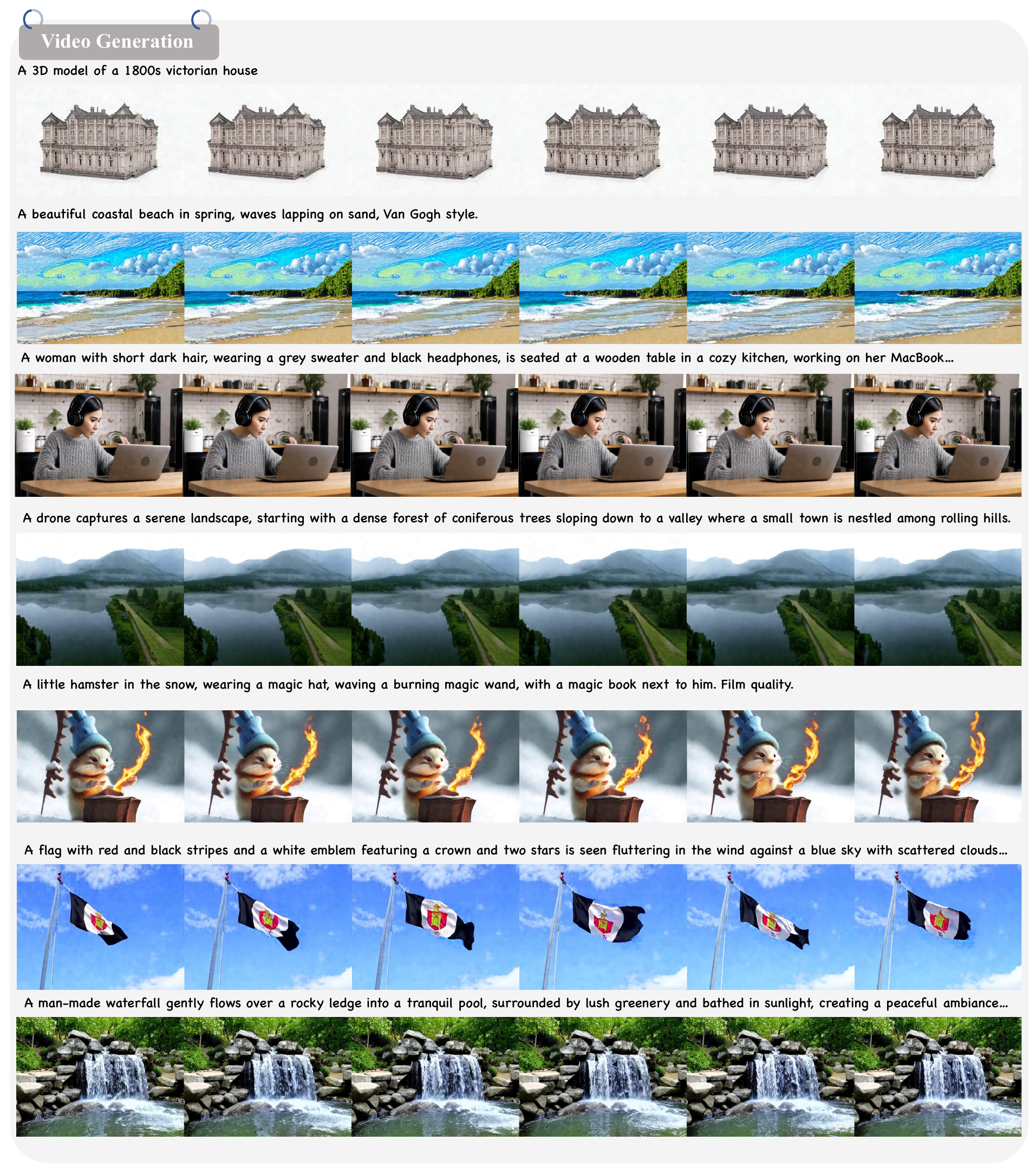}
    \vspace{-3ex}
    \caption{More qualitative results about video generation.}
    \label{fig:more_case}
\vspace{-5ex}
\end{figure*}

\clearpage
{
\bibliographystyle{splncs04}
\bibliography{reference}
}
\end{document}